\documentclass{article}

\usepackage{arxiv}
\usepackage{wrapfig}
\usepackage[utf8]{inputenc} % allow utf-8 input
\usepackage[T1]{fontenc}    % use 8-bit T1 fonts
\usepackage{hyperref}       % hyperlinks
\usepackage{url}            % simple URL typesetting
\usepackage{booktabs}       % professional-quality tables
\usepackage{amsfonts}       % blackboard math symbols
\usepackage{nicefrac}       % compact symbols for 1/2, etc.
\usepackage{microtype}      % microtypography
\usepackage{lipsum}
\usepackage{graphicx}
\usepackage[bottom]{footmisc}
\title{What can the brain teach us about building artificial intelligence?}

\makeatletter
\def\blfootnote{\xdef\@thefnmark{}\@footnotetext}
\makeatother

\author{
  Dileep George
    \\
  Vicarious AI\\
  \texttt{dileep@vicarious.com} \\
}
\vspace{-2cm}
\begin{document}
\maketitle

\begin{abstract}
This paper is the preprint of an invited commentary\footnote{Published in Behavioral and Brain Sciences vol 40, 2017} on Lake et al's Behavioral and Brain Sciences article titled {\em Building machines that learn and think like people} \cite{lake2017building}. Lake et al's paper offers a timely critique on the recent accomplishments in artificial intelligence from the vantage point of human intelligence, and provides insightful suggestions about research directions for building more human-like intelligence. Since we agree with most of the points raised in that paper, we will offer a few points that are complementary.
\end{abstract}

%\blfootnote{Published in Behavioral and Brain Sciences vol 40, 2017}

% keywords can be removed

\section*{Introduction}

\begin{figure}[b]
    \centering
    \includegraphics[width=16.5cm]{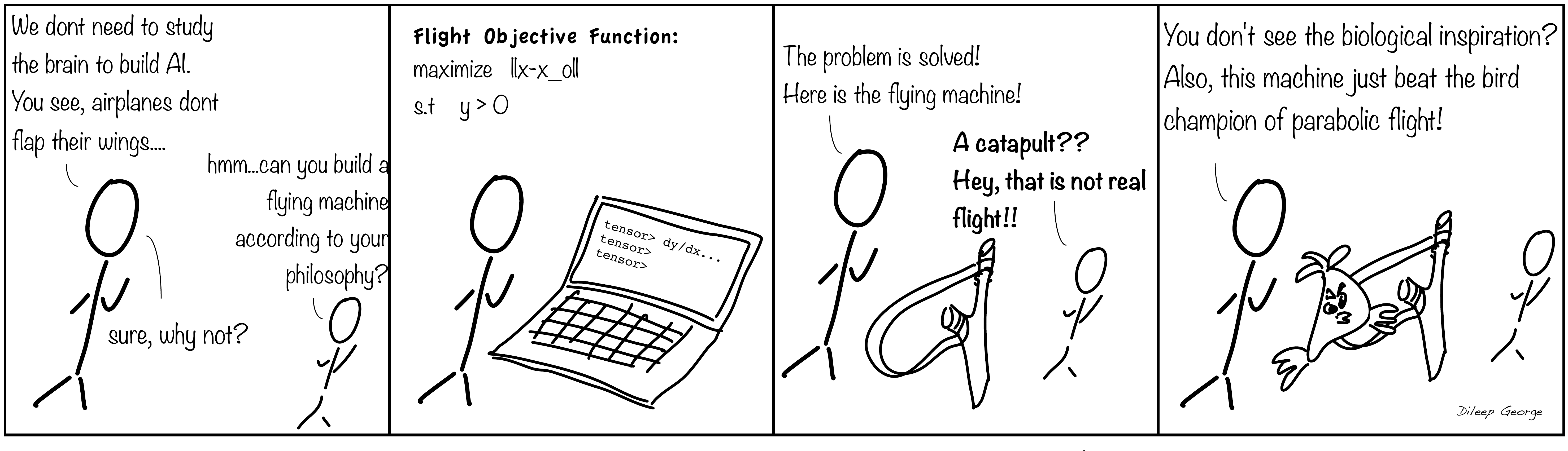}
    \caption{A humorous take on the debate on whether AI can learn anything from the brain.}
    \label{fig:my_label}
\end{figure}

The fact that ‘airplanes do not flap their wings’ is often offered as a reason for not looking to biology for AI insights. This is ironic because the idea that flapping is not required to fly originated from observing eagles soar on thermals! The comic strip in Fig. 1 offers a humorous take on the current debate in AI. A flight researcher who does not take inspiration from birds defines an objective function for flight and ends up creating a catapult. Clearly, a catapult is an extremely useful invention. It can propel objects through the air, and in some cases it can even be a better alternative to flying. Just as researchers who are interested in building ‘real flight’ would be well advised to pay close attention to the differences between catapult flight and the bird flight, researchers who are interested in building ‘human-like intelligence’ or Artificial General Intelligence (AGI), would be well advised to pay attention to the differences between the recent successes of Deep Learning and human intelligence. We believe Lake et al's paper \cite{lake2017building} delivers on that front, and we agree with many of its conclusions.

\section* {Better universal algorithms or more inductive biases?}
\label{sec:headings}
Learning and inference are instances of optimization algorithms. If we could derive a universal optimization algorithm that works well for all data, the learning and inference problems for building AGI would be solved as well. Researchers who work on assumption-free algorithms are pushing the frontier on this question.

Exploiting inductive biases and the structure of the AI problem makes learning and inference more efficient. Our brains show remarkable abilities to perform a wide variety of tasks on data that look very different. What if all these different tasks and data have underlying similarities? That would explain why our brains are remarkably adept at a wide range of problems. Our view is that biological evolution, by trial and error, figured out a set of inductive biases that work well for learning in this world, and the human brain’s efficiency and robustness derives from these biases. When we look to neuroscience for guidance, it should be in search of those goldilock set of assumptions that are specific enough to make learning possible in a short amount of time while general enough to maintain versatility. 

Lake’s paper noted that many researchers hope to overcome the need for inductive biases by bringing biological evolution into the fold of the learning algorithms. We wish to point out that biological evolution had the advantage of using building blocks (proteins, cells) which obeyed the laws of the physics of the world that these organisms were evolving to survive in. In this way, assumptions about the world were implicitly baked into the representations that biological evolution used.  
Trying to evolve intelligence without assumptions will be significantly harder than biological evolution, and it might be infeasible. Even with the help of implicit assumptions, Nature took millions of years and many fortuitous events to evolve human-like intelligence. Therefore it is sensible to take lessons from what it produced.

AGI has one existence proof — our brains. Because of this existence proof, we know that we will one day be able to build AGI. Our brains are general, but that generality has limits. The success of biological evolution in creating our brains is not an existence proof for the same process resulting in an arbitrarily powerful intelligence (call it Artificial Universal Intelligence (AUI)). 
Like perpetual machines, AUI is easy to imagine, but infeasible to build because it depends on physically impossible constructs like infinite computing power or infinite amounts of data. Delineating the limits of generality of human intelligence, and helping to tease apart which of those limits arise from hardware constraints as opposed to fundamental algorithmic limits is another way in which cognitive science and neuroscience can help AI research.

Our strategy is to build systems that strongly exploit inductive biases, while keeping open the possibility that some of those assumptions can be relaxed by advances in optimization algorithms.

\section*{What kind of generative model is the brain? Neuroscience can help, not just cognitive science}
The authors offered several compelling arguments for using cognitive science insights. In addition to cognitive science, neuroscience data can be examined to obtain clues about what kind of generative model the brain implements, and how this model differs from models being developed in the AI community.

For instance, spatial lateral connections between oriented features are a predominant feature of the visual cortex, and are known to play a role in enforcing contour continuity. However, lateral connections are largely ignored in current generative models \cite{lee2015visual}. Another example is the factorization of contours and surfaces. Evidence indicate that contours and surfaces are represented in a factored manner in the visual cortex \cite{zhou2000coding}, potentially giving rise to the ability of humans to imagine and recognize objects with surface appearances that are not prototypical — like a blanket made of bananas, or a banana made of blankets. Similarly, studies on top-down attention demonstrate the ability of the visual cortex to separate out objects even when they are highly overlapping and transparent \cite{cohen2013neural}. These are just a handful of examples from the vast repository of information about cortical representations and inference dynamics, all of which could be used in service of building AGI.

\section*{The conundrum of ‘human-level performance’: benchmarks for AGI}
We would like to emphasize the meaninglessness of ‘human-level performance’, as reported in mainstream AI publications and then used as a yardstick to measure our progress towards AGI. Take the case of the DeepQ network playing ‘breakout’ at a ‘human level’ \cite{mnih2015human}. We found that even simple changes to the visual environment (as insignificant as changing the brightness) dramatically and adversely affects the performance of the algorithm, whereas humans are not affected by such perturbations at all. At this point, it should be well accepted that almost any narrowly defined task can be ‘solved’ with brute force data and computation, and that any use of ‘human-level’ as a comparison should be reserved for benchmarks that adhere to the following principles: — 1) learning from few examples, 2) generalizing to distributions that are different from the training set, and 3) generalizing to new queries (for generative models) and new tasks (in the case of agents interacting with an environment)

\section*{Message-passing (MP) based algorithms for probabilistic models}
While the article makes good arguments in favor of structured probabilistic models, it is surprising that the authors mentioned only MCMC as the primary tool for inference. Although MCMC has asymptotic guarantees, the speed of inference in many cortical areas is more consistent with MP-like algorithms that arrive at maximum a-posteriori solutions using only local computations. Despite lacking theoretical guarantees, MP has been known to work well in many practical cases, and recently we showed that it can be used for learning of compositional features \cite{lazaro2016hierarchical}. There is growing evidence for the use of MP-like inference in cortical areas \cite{bastos2012canonical, george2009towards} and they could offer a happy medium where inference is fast as in neural networks, while retaining MCMC’s capability for answering arbitrary queries on the model.

\section*{Acknowledgments}
I thank Miguel L\'azaro-Gredilla, Scott Phoenix, Nick Hay, and Danny Sawyer for helpful comments on this manuscript. 

\bibliographystyle{unsrt}  
\bibliography{references}  

\end{document}